\documentclass[sigconf]{aamas} 

\usepackage{subcaption}

\usepackage{balance} 

\allowdisplaybreaks

\newcounter{optimization}
\makeatletter
\newenvironment{OptimizationProblem}{%
	\begin{subequations}%
		\refstepcounter{optimization}%
		\addtocounter{parentequation}{-1}%
		\renewcommand {\@currentlabel} {P\theoptimization}%
	} {%
	\end{subequations}
}
\makeatother

\setcopyright{ifaamas}
\acmConference[AAMAS '26]{Proc.\@ of the 25th International Conference
on Autonomous Agents and Multiagent Systems (AAMAS 2026)}{May 25 -- 29, 2026}
{Paphos, Cyprus}{C.~Amato, L.~Dennis, V.~Mascardi, J.~Thangarajah (eds.)}
\copyrightyear{2026}
\acmYear{2026}
\acmDOI{}
\acmPrice{}
\acmISBN{}

\acmSubmissionID{12}

\title{A Multi-Agent system for Multi-Objective constrained optimization}

\author{Federica Filippini}
\affiliation{
  \institution{University of Milano-Bicocca}
  \city{Milan}
  \country{Italy}}
\email{federica.filippini@unimib.it}

\begin{abstract}
Many decision-making problems in computing and networking systems can be naturally formulated as cost-minimization problems under performance constraints. 
In dynamic environments, reinforcement learning (RL) is often used to solve such problems at runtime by embedding both costs and constraint violations into a single scalar reward through weighted penalty terms, following a Lagrangian-inspired formulation. 
However, in this context the behavior of the learned policy critically depends on the choice of these weights, which are typically selected manually. This makes it difficult to identify an appropriate trade-off between optimizing the primary objective and effectively avoiding constraint violations, particularly in non-stationary environments where their relative importance may change. 
This paper presents MAMO (Multi-Agent system for Multi-Objective constrained optimization), an approach to tackle this balancing problem through multi-agent RL. MAMO decouples task execution from objective design by formulating the selection of reward weights as a learning problem, providing a first step towards more autonomous and robust RL-based solutions for constrained optimization problems in dynamic environments.
\end{abstract}


\newcommand{\BibTeX}{\rm B\kern-.05em{\sc i\kern-.025em b}\kern-.08em\TeX}

\begin{document}


\pagestyle{fancy}
\fancyhead{}


\maketitle 


\section{Introduction}\label{sec:intro}

A broad class of decision-making problems in computing and networking systems can be formulated as cost-minimization problems under performance constraints~\cite{ardagna2014}. The objective function typically combines direct costs, such as those incurred when exploiting cloud resources~\cite{filippini2023tsc,li2023,zanussi2024}, and indirect costs, such as energy consumption~\cite{filippini2024sts,sedghani2024tsc} or execution time~\cite{shi2017,zanussi2024}. These problems are subject to quality-of-service (QoS) requirements, including deadline guarantees~\cite{filippini2023tsc,filippini2024sts}, resource budgets~\cite{shi2017}, or throughput constraints. This structure, common across many application domains~\cite{bragin2024,wang2003}, is frequently adopted in resource and application management problems in the computing continuum (CC).

The CC paradigm integrates cloud, edge, fog, and end-device resources to jointly handle latency-sensitive and resource-intensive workloads~\cite{dustar2023}. In this context, tasks such as resource selection, workload scaling, and computation offloading must trade off operational or energy costs against latency and reliability requirements. While static design-time decisions provide useful guidance~\cite{sedghani2024tsc}, rapidly evolving conditions such as fluctuating workloads~\cite{cavadini2024figarov2,filippini2023s4air}, bandwidth~\cite{kambale2025taas}, and resource availability~\cite{talebian2022} require continuous runtime adaptation. 
When system dynamics are complex or partially unknown, classical model-based optimization becomes impractical. As a result, reinforcement learning (RL) is widely adopted to tackle constrained optimization problems at runtime~\cite{cardellini2018,cavadini2024figarov2,kambale2025taas}. 

In these settings, constraints are commonly incorporated into a single scalar reward by aggregating the primary cost with weighted penalty terms for violations, following an approach inspired to Lagrangian relaxation~\cite{cardellini2018,kambale2025taas,bragin2024}. 
This practice results in agents learning to optimize a single objective that implicitly balances efficiency and performance. Despite its conceptual simplicity, this introduces a key limitation: the trade-off between between such conflicting goals is fully determined by manually-chosen weights. Different values can lead to overly conservative or overly aggressive policies, which either prioritize constraint satisfaction at high cost or minimize costs at the price of frequent violations of QoS requirements. Identifying an appropriate balance is difficult and problem-specific, particularly in non-stationary environments where the relative importance of objectives may change over time~\cite{cardellini2018,kambale2025taas}.

To address this limitation, we propose a novel approach called MAMO (Multi-Agent system for Multi-Objective constrained optimization) that decouples task execution from objective design. Its hierarchical architecture integrates a \textit{Task-Execution} (TE) agent that learns a control policy using a standard weighted reward, and a \textit{Weight-Adaptation} (WA) agent that observes long-term system indicators and learns how to adjust the weighting coefficients to align behavior with higher-level objectives. 
MAMO allows the balance between cost efficiency and constraint satisfaction to emerge from experience rather than from hand-tuned parameters. This design enables the system to adapt its notion of ``optimality” as conditions evolve, while preserving the simplicity and compatibility of the standard reward-based RL formulation at the task level.

While the MAMO approach, described in Section~\ref{sec:mamo}, is independent from the specific problem tackled by the TE agent, we propose a preliminary application to a resource scaling problem in edge computing, formulated in Section~\ref{sec:referencescenario} as a cost-minimization problem with QoS constraints. Section~\ref{sec:experiments} presents an experimental analysis highlighting the promising results of MAMO in a simple use-case. Section~\ref{sec:sota} overviews related literature proposals, while Section~\ref{sec:conclusion} discusses conclusions and plans for future research.


\section{Reference Problem}\label{sec:referencescenario}

\begin{figure}[t]
    \centering
    \includegraphics[width=.7\columnwidth]{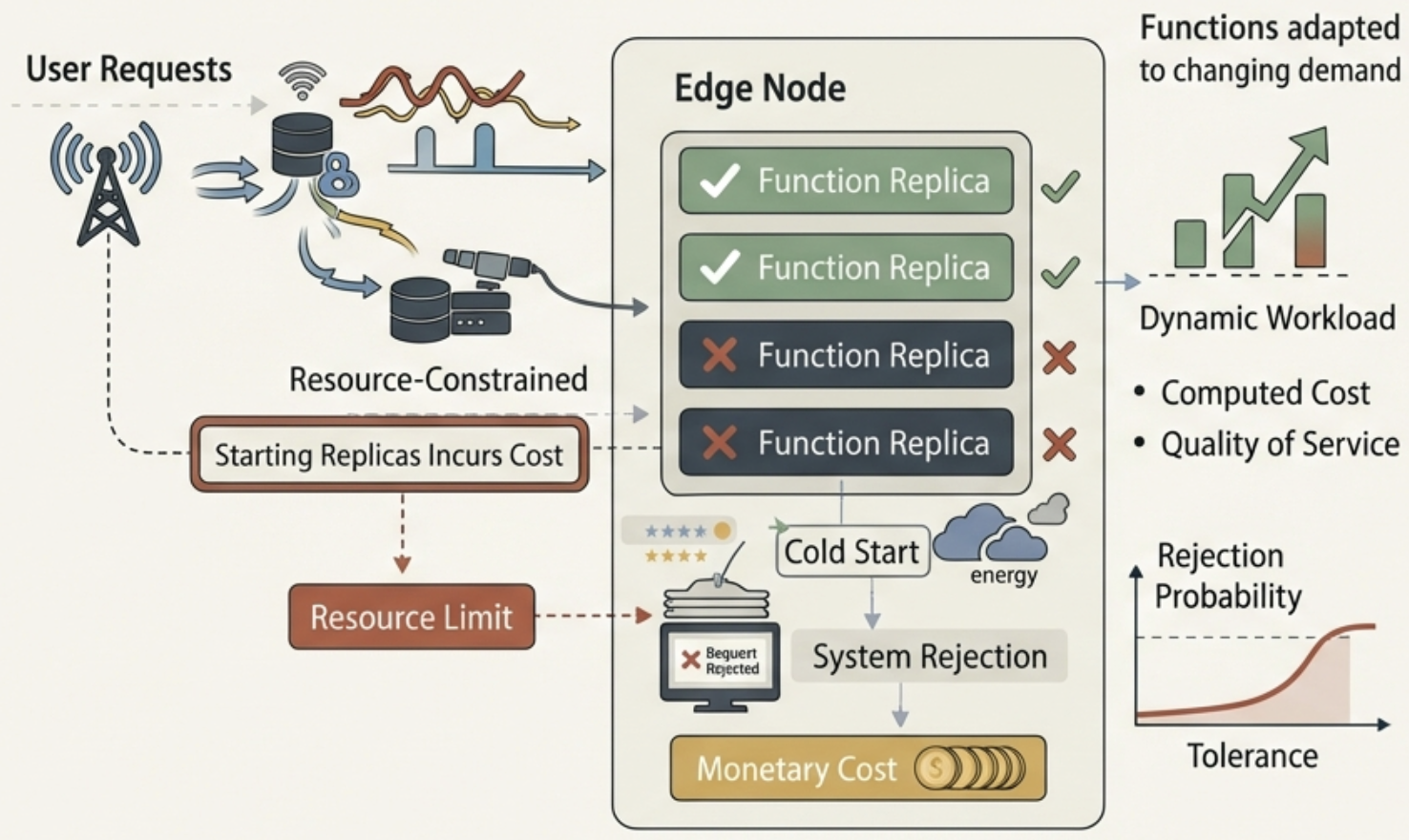}
    \caption{The replica scaling problem in edge-FaaS systems}
    \label{fig:scaling}
\end{figure}

To illustrate the characteristics of the class of problems addressed in this work, we consider the replica scaling problem in a Function-as-a-Service (FaaS)-enabled edge computing environment. FaaS platforms allow applications to be decomposed into stateless functions that are instantiated on demand and executed in response to incoming requests. To sustain a given workload while meeting response-time requirements, multiple replicas of the same function can be deployed concurrently on an edge node (see Figure~\ref{fig:scaling}).

FaaS workloads are typically highly dynamic, driven by user mobility, time-of-day effects, and context-dependent application behavior~\cite{tundo2025tnsm}. If too few replicas are provisioned, incoming requests may queue up, leading to increased response times, and eventually rejected by the system due to reaching the maximum capacity~\cite{mahmoudi2020tccserverless}. At the same time, however, while resources in cloud data centers are virtually unlimited, edge nodes are inherently resource-constrained. The available memory and compute capacity limit the number of function replicas that can be instantiated concurrently. This makes the scaling problem at the edge fundamentally different from its cloud counterpart: while over-provisioning is often feasible in centralized infrastructures, at the edge excessive replication can quickly lead to resource exhaustion and performance degradation. As a result, the number of active replicas must be carefully selected to balance responsiveness and resource efficiency.

In addition to resource constraints, scaling actions themselves incur non-negligible costs. Starting a new function replica may involve container initialization, memory allocation, and code loading, which translate into latency and resource overheads commonly referred to as cold-start costs~\cite{wang2025optimizing}. In commercial settings where resources are offered on a pay-per-use basis, these overheads may also correspond to direct monetary costs, while in private or energy-aware deployments they can be interpreted as energy consumption or as a proxy metric capturing the operational burden of frequent scaling actions. These costs must be accounted for together with performance objectives when deciding how to adapt the system.

Formally, if we denote by $\mathcal{F} = \{1,\dots,F\}$ the set of functions $f$ that can be deployed on an edge node, and we introduce integer decision variables $n_f$ that represent the number of replicas to be initialized for function $f$ according to the value of incoming workload $\lambda_f$ (i.e., the expected rate of requests per second that reach the node in a pre-defined control period), we can express the scaling problem as: 

\begin{OptimizationProblem}
    \begin{equation}
        \min \sum_{f\in\mathcal{F}}c_fn_f  \label{eq:objective}
    \end{equation}

    subject to:
    \begin{align}
        0 \leq n_f\leq N_f & \qquad \forall~f\in\mathcal{F}\\
        p(\lambda_f,n_f) \leq tol& \qquad \forall~f\in\mathcal{F},\label{eq:rejectprob}
    \end{align}
    \label{opt:scaling}
\end{OptimizationProblem}

where $c_f$ represents the cost associated with initializing a replica of function $f$, $N_f$ denotes the maximum number of available replicas, and $p(\lambda_f,n_f)$ is the probability of experiencing requests rejections given the expected incoming load $\lambda_f$ and the selected number of replicas (which should be lower than a tolerance to ensure system responsiveness). 
In particular, $N_f$ can be a static value imposed within the FaaS platform (e.g., OpenFaaS\footnote{\url{https://docs.openfaas.com/architecture/autoscaling}}) or a dynamic value that depends on the function memory and resource requirements and on the node current operating conditions. 
Similarly, the probability $p(\lambda_f,n_f)$ can be estimated through mathematical models based on, e.g., queuing theory~\cite{mahmoudi2020tccserverless} or predicted based on historical data.

In practice, the parameters that characterize this problem are inherently non-stationary: fluctuations of the incoming workload come with varying resource availability at the edge node due to co-located services or background processes, while the performance of replicas may change as a result of contention and network dynamics. Accurately modeling the joint effect of these factors on the rejection probability $p(\lambda_f,n_f)$ is challenging, as it depends on complex and time-varying interactions that are difficult to capture with static analytical models. As a consequence, solutions based on fixed assumptions or offline optimization rapidly become suboptimal as operating conditions evolve, making the replica scaling problem particularly well-suited to RL-based approaches. 

As mentioned in Section~\ref{sec:intro}, mapping the replica scaling problem to the RL setting requires to express Constraints~\eqref{eq:rejectprob} as part of the reward that the environment provides to the agent to evaluate the quality of the selected action. Indeed, RL agents iteratively develop a \textit{policy}, i.e., a function that maps the current \textit{state} $s$ of the environment to an \textit{action} $a$, which impacts on the transition to a new state $s^\prime$. The policy is continuously updated to maximize the expected return $\sum_{t=1}^\infty\gamma^tr(s,a,s^\prime)$, where $\gamma$ is a parameter that discounts the expected reward $r$ in
future time steps, thus limiting the impact of distant choices on the
learned policy.

In this context, the state $s$ observed by the agent encodes information on the current system conditions (e.g., the incoming workload $\lambda_f$), while the action $a$ consists in selecting an appropriate number of replicas $n_f$ for each function $f$. As already discussed, we can define $r(s,a,s^\prime) = 1 - C(s,a,s^\prime)$, where $C(s,a,s^\prime)$ is a cost function that incorporates the replica instantiation cost and a penalty for high rejection rates, e.g., $C(s,a,s^\prime) = \sum\limits_{f\in\mathcal{F}}\left(w^0_fc_f\frac{n_f}{N_f} + w^1_f p(\lambda_f,n_f)\right)$. As in~\cite{cardellini2018,kambale2025taas}, the weights $w^i_f\in[0,1]$ for a given function $f$ quantify the relative importance of limiting its execution costs against avoiding constraints violations; considering different functions, instead, $w^1_{f}$ and $w^0_f$ identify the function priority. We have $\sum_{i,f}w^i_f = 1$.

\section{The MAMO Approach}\label{sec:mamo}

\begin{figure}[t]
    \centering
    \includegraphics[width=.8\linewidth]{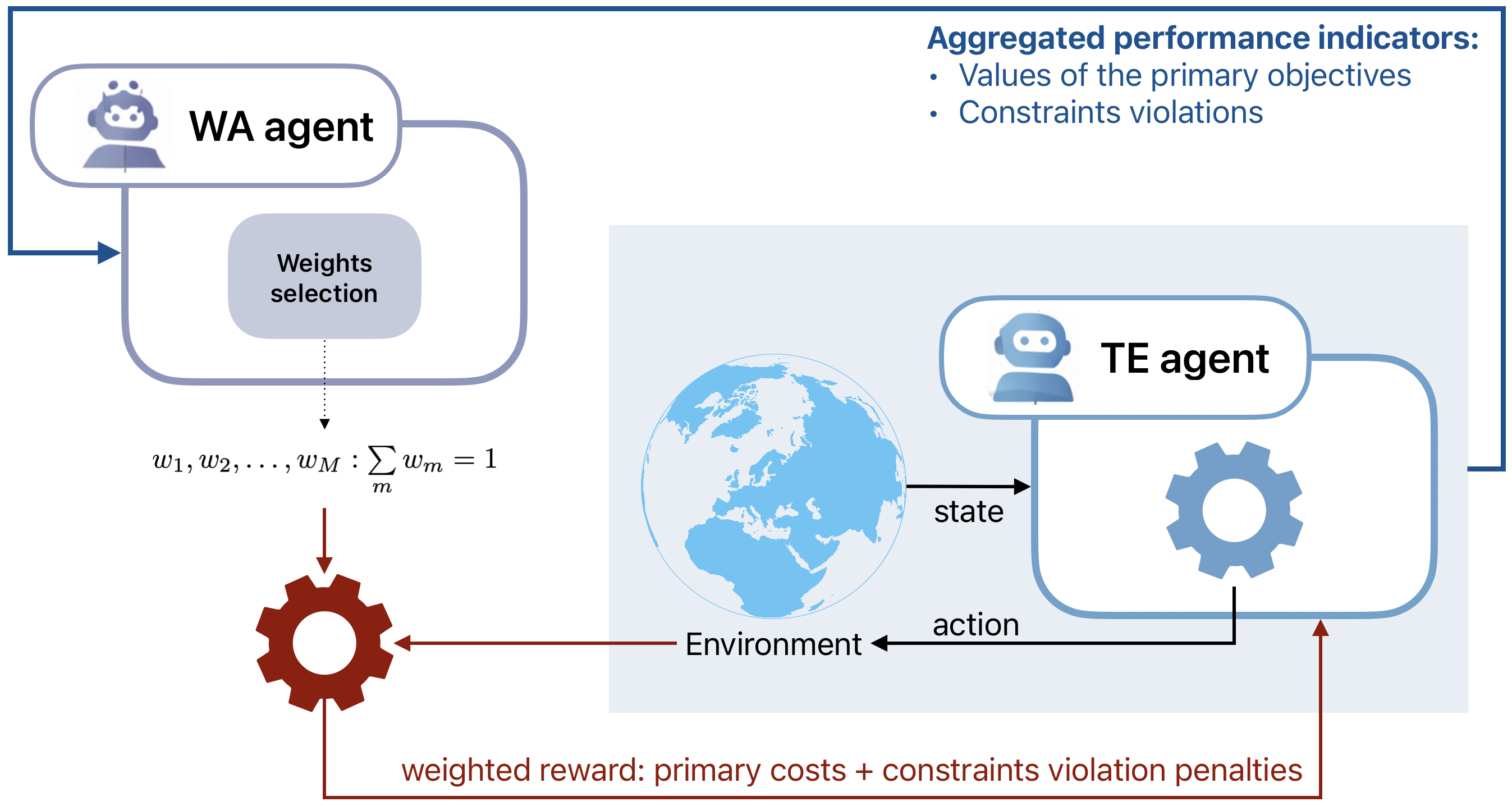}
    \caption{MAMO architecture}
    \label{fig:mamo}
\end{figure}

MAMO (Multi-Agent system for Multi-Objective optimization) is a hierarchical multi-agent framework designed to learn the trade-off between conflicting objectives in constrained cost-minimization problems. MAMO explicitly separates task execution from objective design by introducing two learning entities that operate at different time scales and abstraction levels, treating the selection of the reward-weighting coefficients as a learning problem rather than as a fixed design choice. To this end, the framework is composed of two agents with complementary roles, as shown in Figure~\ref{fig:mamo}:

\begin{itemize}
    \item A Task-Execution (TE) agent that interacts directly with the environment and learns a control policy using a standard reward that aggregates the primary cost and the constraint-violation penalties through a weighted sum. For fixed values of the weights, the TE agent behaves as a conventional RL agent, optimizing the corresponding composite objective.

    \item A Weight-Adaptation (WA) agent that operates at a higher level and at a slower time scale. It does not act on the environment directly, but instead selects the values of the weights that regulate the trade-off between the conflicting objectives in the TE reward function, identifying through experience the weighting configuration that best aligns the TE agent behavior with the desired system-level objectives.
\end{itemize}

MAMO adopts a two-phases iterative workflow where: (1) a weight is selected by the WA agent and fixed for a given training horizon; during this phase, the TE agent interacts with the environment and updates its policy to optimize the corresponding weighted reward. (2) At the end of the learning phase, the WA agent observes aggregated performance indicators produced by the TE agent and evaluates the quality of the selected weight. Based on this feedback, it selects a new weight, and a new learning phase for the TE agent is started. 
Through this loop, the WA agent progressively learns how to steer the TE agent by shaping its reward, rather than by directly controlling its actions. In this way, MAMO decouples how a task is solved from which trade-off should be enforced.

In the reference scenario described in Section~\ref{sec:referencescenario}, the TE agent is responsible for selecting, at each control period, the number of replicas $n_f$ for each function $f \in \mathcal{F}$. Its reward is defined as in the previous section, where weights $w^i_f$ regulate the balance between the replica instantiation cost and the rejection probability. 

In this setting, the WA agent controls the values of $w^i_f$. At the beginning of each adaptation cycle, candidate values are selected (initially at random) and provided to the TE agent. At the end of the TE training phase, the system performance is summarized through two indicators, which correspond to the state observed by the WA agent: the average execution cost incurred by the selected replicas, and the average probability of request rejection. 
This also receives a scalar reward accordingly: if the rejection probability exceeds the tolerance threshold defined in Constraints~\eqref{eq:rejectprob}, the reward is set to zero, thus discouraging weight selections that lead to constraint violations. Otherwise, the reward is equal to the execution cost, so that lower-cost configurations are preferred when QoS requirements are satisfied\footnote{While in the original optimization problem the tolerance is imposed on the rejection probability of each individual function, in our MAMO implementation it is enforced on average over the observation window; although this check is less restrictive, it still provides a meaningful high-level indicator of the TE agent overall performance.}. 
Formally, the WA agent therefore learns a policy that maps observed performance summaries to a new value of the weights $w^i_f$, with the objective of minimizing execution costs while enforcing the rejection probability constraint. The new weights are then passed to the TE agent, and the learning loop is repeated.

This hierarchical interaction allows MAMO to progressively refine the trade-off between cost efficiency and QoS preservation directly from experience, without relying on manual tuning or fixed penalty coefficients. As a result, the system can adapt its notion of optimality as workload patterns and resource conditions evolve, while retaining the simplicity of a standard reward-based RL formulation at the task-execution level.

\section{Experimental Analysis}\label{sec:experiments}

\begin{figure}[t]
\centering
\begin{subfigure}[t]{.25\textwidth}
    \centering
    \includegraphics[width=\linewidth]{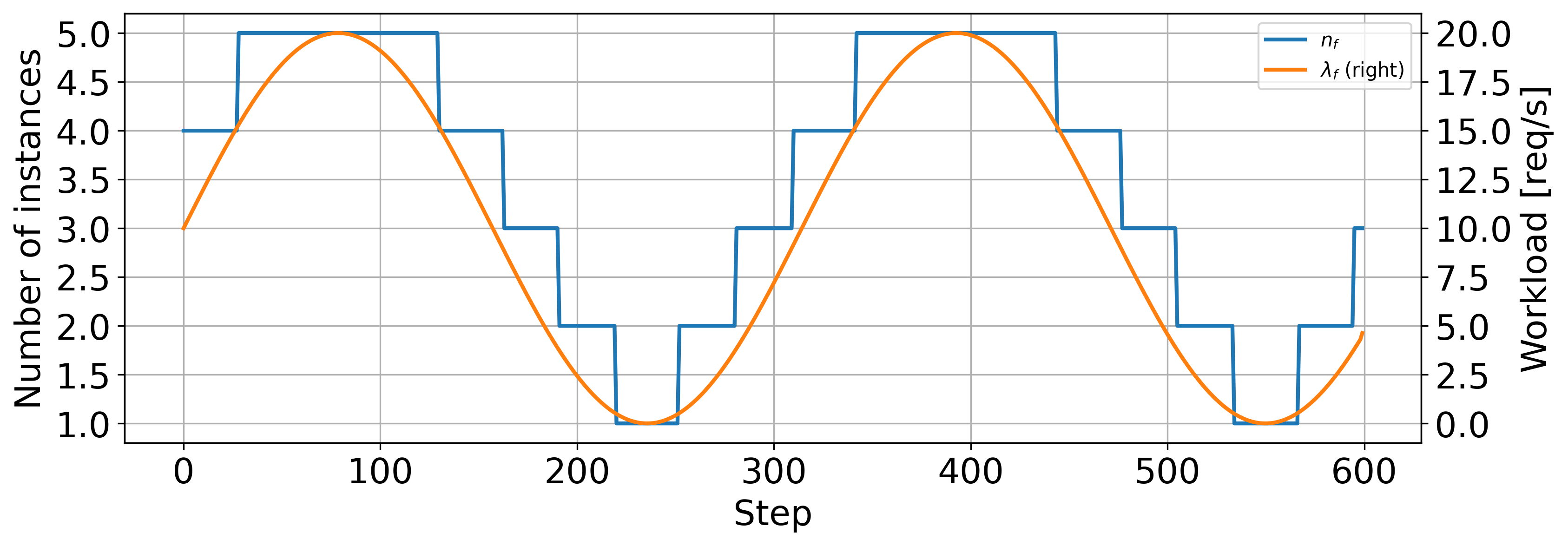}
    \vfill
    \includegraphics[width=\linewidth]{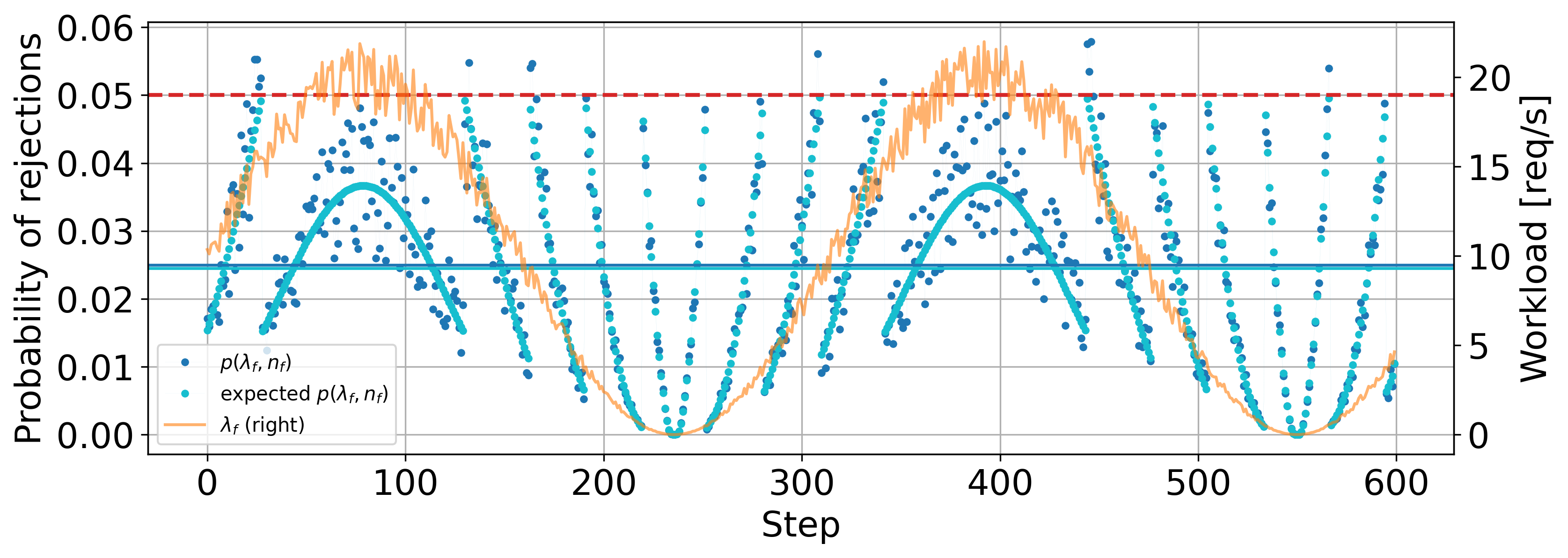}
    \end{subfigure}
    \caption{$n_f$ and $p(\lambda_f,n_f)$ observed solving Problem~\eqref{opt:scaling}}
    \label{fig:model_sol}
\end{figure}

\begin{figure}[t]
    \centering
    \includegraphics[width=.47\linewidth]{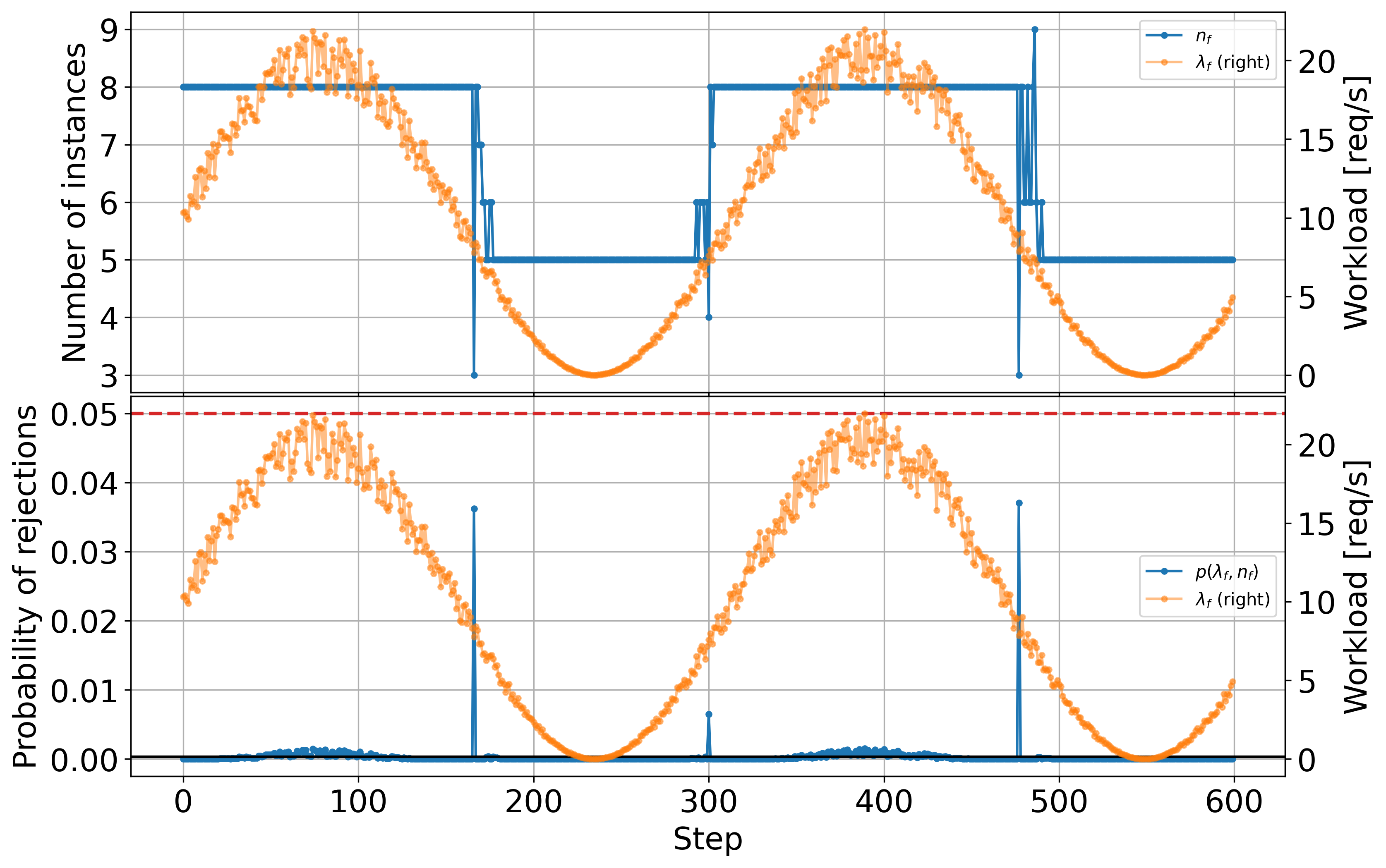}
    \hfill
    \includegraphics[width=.47\linewidth]{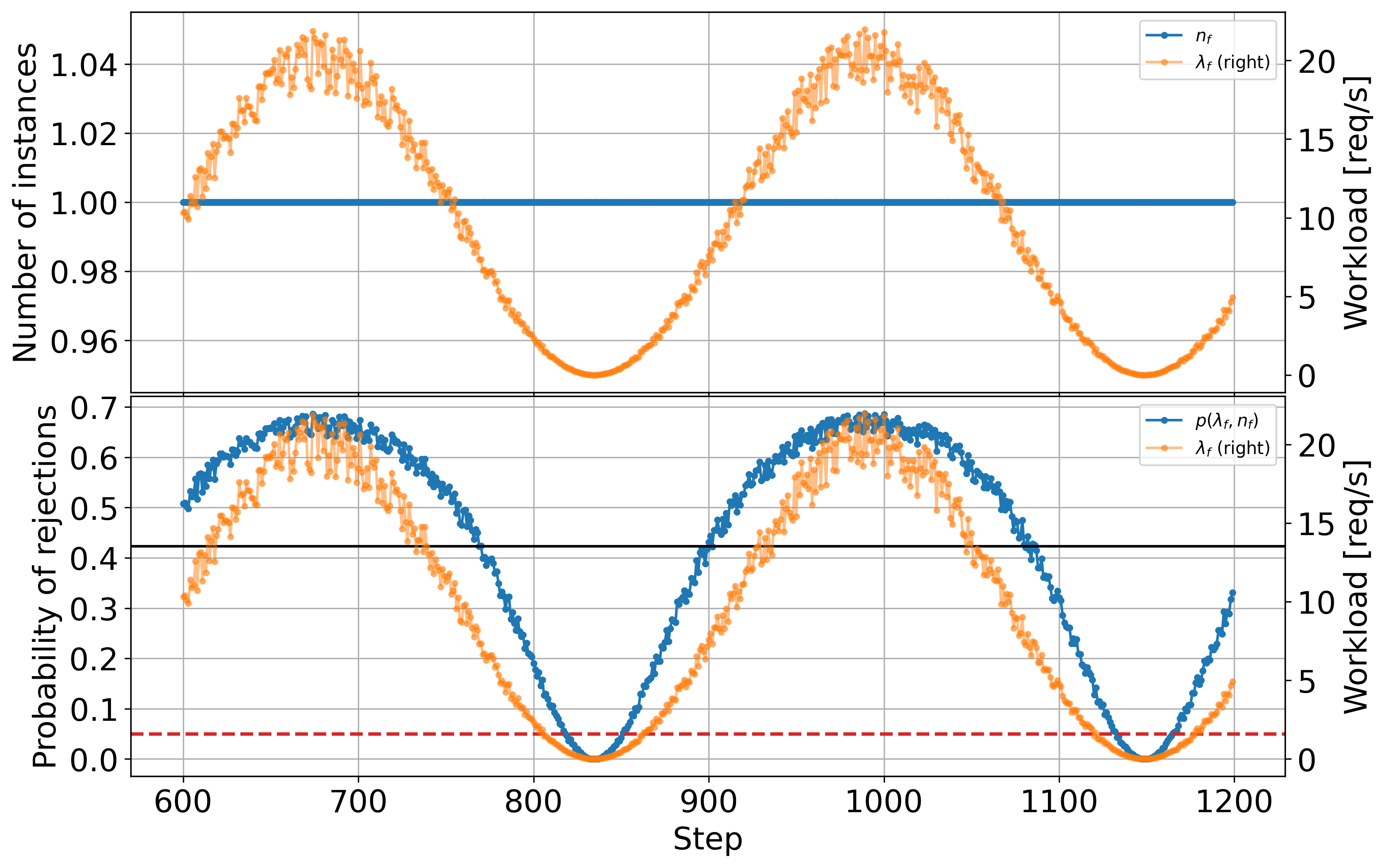}
    \caption{$n_f$ and $p(\lambda_f,n_f)$ after training TE with $w = 0.99$ (left) and $w=0.1$ (right); evaluation lasts $600$ steps}
    \label{fig:res_099}
\end{figure}

We evaluate MAMO in a simplified instance of the reference scenario from Section~\ref{sec:referencescenario}, by considering a single function $f$ (and, thus, a single $w$ that multiplies $p(\lambda_f,n_f)$, while $1-w$ multiplies the cost). This allows to isolate the effect of the weight-adaptation mechanism and observe the trade-off between execution costs and constraint violations. 
The workload $\lambda_f$ follows a sinusoidal trace, as adopted in the literature to emulate diurnal patterns~\cite{petriglia2025}. This model captures non-stationary dynamics while remaining analytically tractable. 
The environment simulates a FaaS edge node, where the TE agent controls the number of replicas $n_f$ of the single function at each control step, considering $N_f = 10$. This is in line with the $N_f = 5$ imposed by OpenFaaS community edition\footnote{\url{https://docs.openfaas.com/architecture/autoscaling/\#legacy-scaling-for-the-community-edition-ce}}, while other $5$ replicas were added to enlarge the TE action space. Cold and warm execution times are set to $1.0 s$ and $0.1 s$~\cite{wang2025optimizing}, respectively, and an idle replica is terminated after $60 s$ of inactivity. We: 

\begin{enumerate}
    \item Solved Problem~\eqref{opt:scaling} offline using Gurobi optimizer 12.0.2\footnote{\url{https://support.gurobi.com/hc/en-us}} for each $\lambda_f$ value, setting $tol = 0.05$, to identify the best achievable performance under perfect knowledge of $\lambda_f$ (see Figure~\ref{fig:model_sol}; results are always reported on an evaluation trace of $600$ steps). This provides a lower bound on the $n_f$ value (and, thus, the execution cost) that keeps $p(\lambda_f,n_f)<tol$.
    \item Perturbed the workload traces multiplying $\lambda_f$ by a uniform noise between $0.9$ and $1.1$ evaluated the solution computed with~\eqref{opt:scaling} considering the expected $\lambda_f$, observing the actual $p(\lambda_f,n_f)$ under the perturbed load (see Figure~\ref{fig:model_sol}, below); while average performance are maintained, adaptation is needed in this context to avoid violations.
    \item Trained and evaluated the TE agent with $w=0.99$ and $0.1$ under the noisy load trace to demonstrate the extreme behaviors resulting from such choices (see Figure~\ref{fig:res_099}).
    \item Run the full MAMO framework: the system is initialized with $w=0.99$; the TE training phase lasts $15k$ iterations under the current weight. The WA agent observes the average probability of rejections $\overline{p}$ and the average costs over the last $300$ steps, and receives a positive reward if $\overline{p}<0.05$.
\end{enumerate}

\begin{figure}[t]
    \centering
    \includegraphics[width=.7\linewidth]{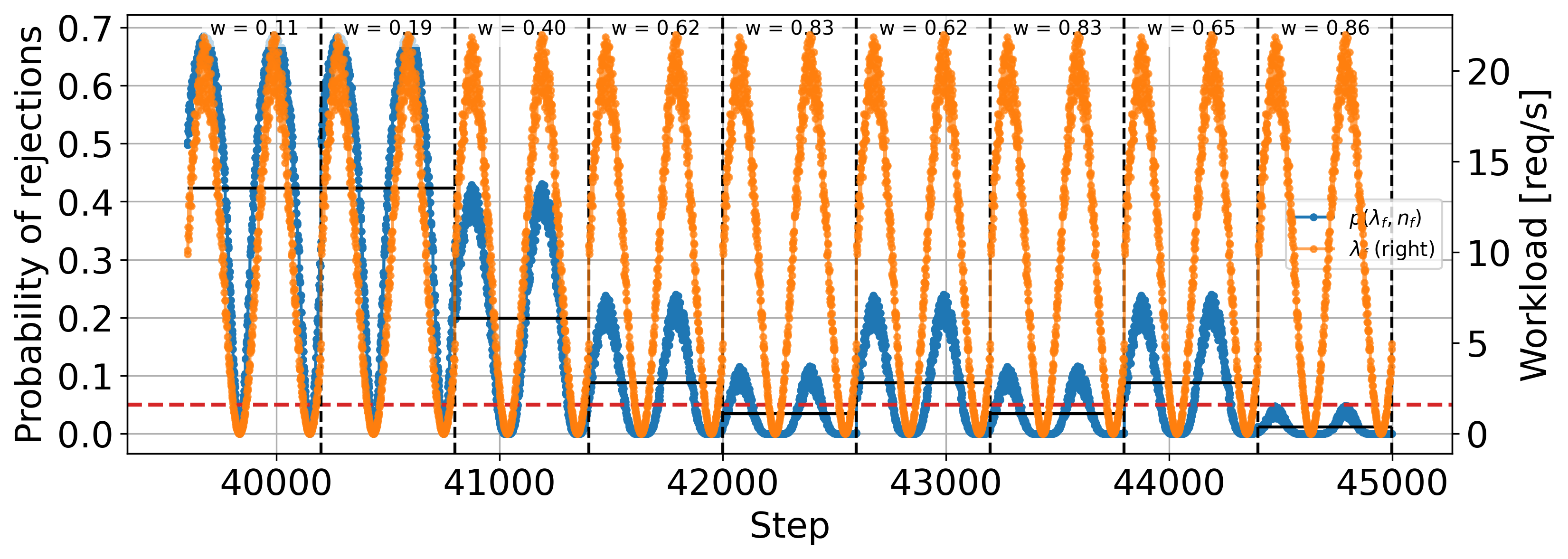}
    \hfill
    \includegraphics[width=.7\linewidth]{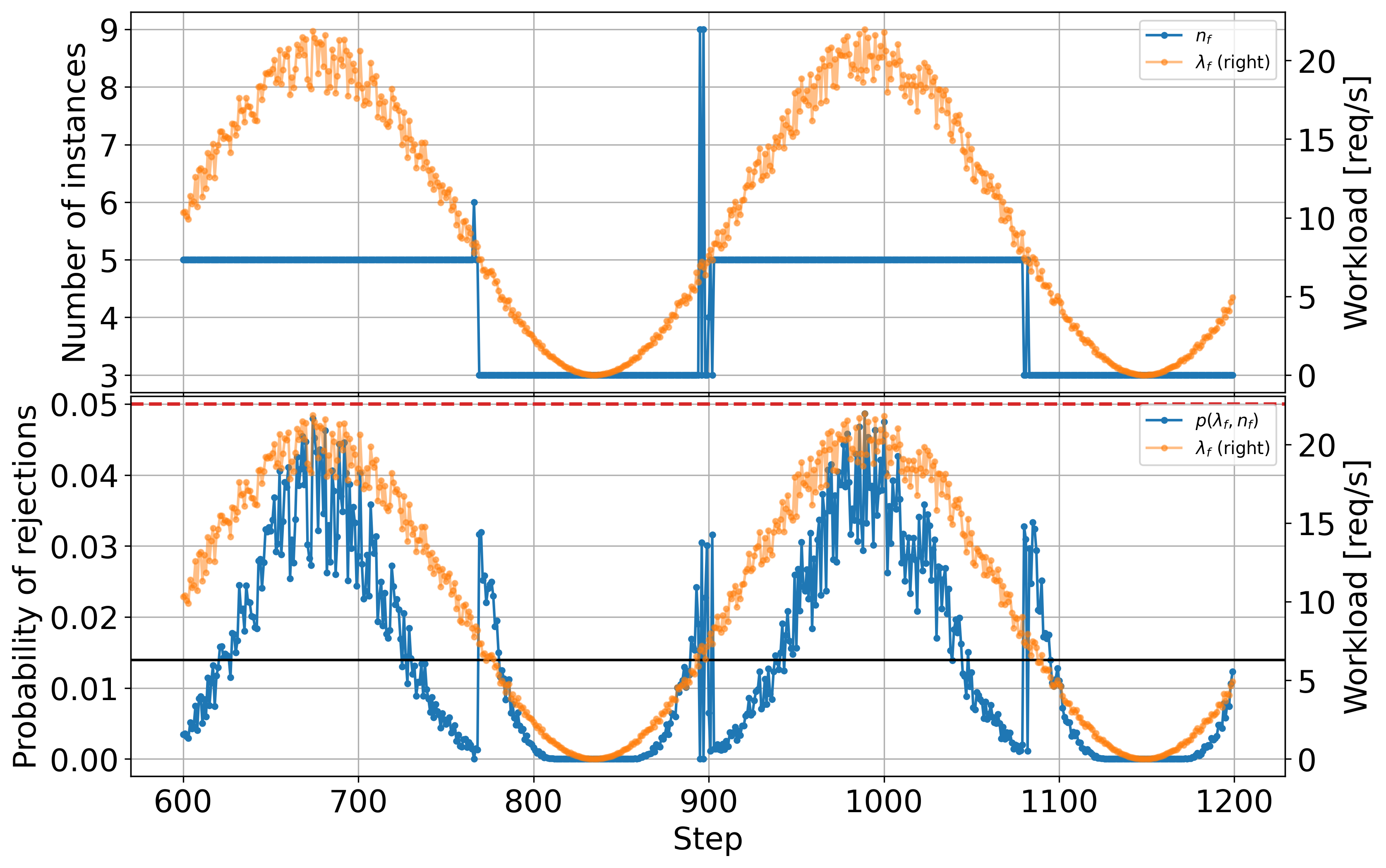}
    \caption{$p(\lambda_f,n_f)$ while the WA agent training proceeds (above) and detailed solution ($n_f$ and $p(\lambda_f,n_f)$) in a representative scenario with $w = 0.85$ (middle and below)}
    \label{fig:mamo_rejects}
\end{figure}

Both agents are trained using RL4CC\footnote{\url{https://github.com/FFede0/RL4CC}}, an open-source library that extends Ray RLlib\footnote{\url{https://docs.ray.io/en/latest/rllib/index.html}}. They are implemented using Deep Q-Learning with a three-layer fully connected network $[256, 128, 256]$, discount factor $\gamma=0.7$, learning rate $5\times10^{-4}$, and a prioritized replay buffer with capacity $10240$. Target networks are updated every $1000$ steps. Exploration follows an $\varepsilon$-greedy strategy with a piecewise schedule synchronized with the $15k$-iteration MAMO cycles. The WA action space is discretized with a step of $0.01$.

We observe from Figure~\ref{fig:mamo_rejects} that, while the WA agent training proceeds, the value of $\overline{p}$ observed when evaluating the TE agent (black horizontal lines) approaches $0.05$, while $w$ converges to values between $0.8$ and $0.9$. Although $n_f$ (and thus the execution cost) is slightly higher than in Figure~\ref{fig:model_sol}, the learned policy is able to adapt to the noisy workload and consistently keep $p(\lambda_f,n_f)$ below $0.05$, highlighting the promising performance of the MAMO approach.

\section{Related Works}\label{sec:sota}

Research on agents that adapt their own learning objectives spans several lines of work. 
Among these, optimal reward frameworks~\cite{gupta2023,singh2010,sorg2010,liu2014} treat the internal reward function as a design variable to be optimized for long-term performance under a fixed external objective. Like MAMO, they recognize that the learning signal need not coincide with the task reward. However, they typically rely on rich, differentiable reward parameterizations and gradient-based optimization, whereas MAMO restricts adaptation to a low-dimensional vector of scalarization weights over pre-defined objectives, optimized by a separate agent.

Meta-learning approaches~\cite{zheng2018,pappalardo2024,jaderberg2019,sheikh2022} similarly adapt training signals using meta-gradients, neural networks, or population-based methods. Compared to these, MAMO adopts a more structured and interpretable form of objective adaptation: the WA agent controls only a global weight vector in a linear combination of cost and constraint penalties, rather than a free-form intrinsic reward model, and explicitly targets a constrained, multi-objective setting.

Meta-gradient~\cite{xu2018} and bi-level~\cite{hu2020,gupta2023} RL also frame reward design as a nested optimization problem, but typically require differentiability of the inner learner and learn fine-grained, state-dependent reward transformations. In contrast, MAMO uses a model-free outer loop that does not differentiate through the TE agent and holds the weight vector fixed over a training horizon, making it compatible with arbitrary inner solvers at the cost of coarser control.

Multi-objective RL addresses vector-valued rewards and conflicting objectives~\cite{hayes2022}, often through weighted scalarization and policies conditioned on preference vectors~\cite{abels2019,chen2019,lu2025}. In most cases, however, these weights are exogenous or optimized by embedded gradient procedures. MAMO instead casts the choice of weights as a sequential decision problem, introducing an explicit WA agent whose action space is the space of trade-offs. 
Finally, while intrinsic-reward learning in multi-agent RL~\cite{liu2014,yang2023} also modifies reward signals to improve coordination, MAMO differs in that its WA agent does not act in the environment and relies only on global performance summaries, positioning it as a supervisory mechanism for objective design rather than a decentralized shaping method.

\section{Conclusion}\label{sec:conclusion}

This paper introduced MAMO, a hierarchical multi-agent framework that learns to balance conflicting objectives in constrained optimization problems by adapting reward weights at runtime. The experimental results in a dynamic edge-FaaS scaling scenario highlight its promising performance: MAMO autonomously converges to weight configurations that satisfy QoS constraints while achieving reasonable execution costs. 
As future work, we plan to evaluate MAMO on problems from different application domains and to compare it with weight-selection strategies such as dual‑decomposition
schemes, Bayesian optimization and multi-policy algorithms (e.g., Optimistic Linear Support), to further assess its effectiveness.



\bibliographystyle{ACM-Reference-Format} 
\bibliography{bibliography}


\end{document}